# Reclassification formula that provides to surpass *K*–means method

M. Kharinov

St. Petersburg Institute for Informatics and Automation of Russian Academy of Sciences (SPIIRAS)

khar@iias.spb.su

**Abstract**

The paper presents a formula for the reclassification of multidimensional data points (columns of real numbers, "objects", "vectors", etc.). This formula describes the change in the total squared error caused by reclassification of data points from one cluster into another and prompts the way to calculate the sequence of optimal partitions, which are characterized by a minimum value of the total squared error $E$ (weighted sum of within–class variance, within–cluster sum of squares WCSS etc.), i.e. the sum of squared distances from each data point to its cluster center. At that source data points are treated with repetitions allowed, and resulting clusters from different partitions, in general case, overlap each other. The final partitions are characterized by "equilibrium" stability with respect to the reclassification of the data points, where the term "stability" means that any prescribed reclassification of data points does not increase the total squared error $E$.

It is important that conventional *K*–means method, in general case, provides generation of instable partitions with overstated values of the total squared error $E$. The proposed method, based on the formula of reclassification, is more efficient than *K*–means method owing to converting of any partition into stable one, as well as involving into the process of reclassification of certain sets of data points, in contrast to the classification of individual data points according to *K*–means method.

**Keywords:** total squared error, minimization, overlapping partitions, *K*–means method, reclassification formula, stability condition.

## 1. INTRODUCTION

In this paper we present in terms of abstract multidimensional cluster analysis, the generalized formulas [1], recently implemented for segmentation of visually perceived images.

In the field of image processing so–called unsupervised image segmentation can be treated as clustering of multidimensional image points (pixels). So, in the context of clustering we prefer to replace the term "pixel" with the term "data point", avoiding mixing of concepts. The term "object" seems inapposite for us, as in image processing terminology this intuitive term appeals to the visual perception and is reserved to designate the desired cluster of data points that we intend to calculate. We prefer to use the term "data point" instead of the term "vector" since "repeatable data point" sounds better than "repeatable vector."

## 2. PROBLEM STATEMENT

The clustering problem is treated in the mathematical formulation [2]. In this case, the required partitions of the data points into clusters are characterized by optimum quality. One of the classic partition quality assessments is the total squared error $E$ to be minimized for each possible number of clusters.

At least two trivial optimal partitioning of data points are known beforehand. These are the partition consisting of a single cluster, and also the partition of individual data points or clusters of identical data points, if the latter is not limited by other conditions. Thus, the problem is reduced to calculating of the remaining optimal partitions that are obtainable by means of iterative transformations of known ones in split–and–merge clustering techniques, combined with proper cluster corrections to provide the generation of overlapping partitions.

It should be noted that, as a rule, for the analysis of clustering results, the initial members from the sequence of partitions of data points into 2, 3, 4, ... clusters are most interesting. But just these partitions obtained by means of enlargement of clusters, turn out dependent on the variation of the starting partitioning of data set into clusters of one or several data points, due to accumulation of modifications of partitions at multiple iterations. Obviously, in the case of proper minimization of the total squared error $E$, this shortage is overcome, since the optimal partitions are defined for different number of clusters independently of each other, and in the outlined framework, the clustering problem seems rather computational than theoretical.

## 3. STATE OF THE ART

The partition quality assessment by means of the total squared error $E$ is used in versions of $K$–means method [2-6], in Otsu method [7, 8] and, with certain reservations, in Mumford–Shah model [9-12][1].

Conventional $K$–means method [3, 4], that protected from algorithmic infinite looping according to [5] and adopted to obtain a sequence of overlapping partitions [6] owing to increment of cluster number, is extensively used in a number of subject areas, including processing of digital images. Otsu method in original [7] and multi–threshold version [8], as well as conventional [9, 10] and modern [11, 12] versions of Mumford–Shah model, based on the famous formula for the criterion of cluster merging, reasonably proved themselves in a variety of applications to early processing of image data. However, the effectiveness of these methods is proved only in technical aspects, e.g. in the field of image processing they are intensively used for detection of visually perceived "objects", which, however, are often lost in the automatic search. To refine the discussed techniques at the expense of reliable minimization of objective functional, their analytical verification and accurate interpretation are still required. It is relevant, especially as the core developers do not overestimate the abilities of their algorithms always to converge to the desired global and even "local" minima [5, 11].

In $K$–means method, the clusters crumbles into individual points, which are then grouping around the nearest centers of previously computed clusters. The closest distance to the cluster centers is sufficient, but not always necessarily condition for optimal clustering. Therefore, it is not guaranteed that the partition obtained by this method can't be improved by means of reclassification of several or even one data point from the one cluster to another. Thus, the main drawback of the $K$–means method is the utilization of intermediary $k$ mean values themselves that are treated as authorized representatives of original sets to update the classification.

The principal limitation of Mumford–Shah model for those or other functional [9-12] is caused by that just only cluster merging is analytically described and is treated alone[2]. So, the analytical generalization and development of the formulas [9-12] for cluster merging is appropriate.

In contrast to the above other methods, the one–dimensional multi–threshold Otsu method [8] performs minimization step by step in strict accordance with the mathematical formulation of the problem. But it faces with an exponential decrease in computational speed and provides the obtaining of only several optimal partitions instead of all of partitions, as it is required. It is remarkable that one–dimensional Otsu method [7, 8] accounts for "compactness" (or "continuity") of optimal partitions, that consists in that a number of clusters is equal to a number of value ranges, occupied by a series of different values of data points. In this case, an exhaustive search of thresholds for values of data points provides to avoid of exhaustive search of partitions. So, the task of unconditional search for optimal partitions is reduced to further optimization of computations without significant deterioration of partition quality assessment.

## 4. RECLASSIFICATION FORMULA

Let $I_1$ and $I_2$ be two central points of clusters 1 and 2, respectively. Let $I$ be the centre of $k$ data points from cluster 1, calculated as mean point for $k$ specified ones. Let $n_1$ be the number of data points in the cluster 1 and $n_2$ be the number of data points in the cluster 2.

Let's reclassify the specified $k$ data points from the cluster 1 into the cluster 2. Then the increment $\Delta E$ in the total squared error $E$, caused

---

[1] Version [11] of Mumford–Shah model, which, in addition to the total squared error $E$, takes into account the total length of the boundaries between adjacent sets of data points, is most often used in practice of image processing. However, according to our experience, the utilization in the version [12] of only classic assessment $E$ of the quality of partitions, regardless of mentioned boundaries, is no less effective.

[2] Less important is the fact that so–called a "regularization parameter" $\lambda$ is permanently varied in the computational process [11], thus breaking the idea of prescribed functional minimizing at all.
An optional condition of connectivity of data points within the clusters [9-12] is also one of less important.

by this reclassification, is given by the following reclassification formula:

$$\Delta E = \begin{cases} \Delta E_{merge}, & k = n_1, \\ \Delta E_{correct}, & k < n_1, \end{cases} \quad (1)$$

where $\Delta E_{merge}$ is the merging criterion [12] given by the expression:

$$\Delta E_{merge} = \frac{\|I_1 - I_2\|^2}{\frac{1}{n_1} + \frac{1}{n_2}} \quad (2)$$

and $\Delta E_{correct}$ is expressed by the formula:

$$\Delta E_{correct} = \frac{\|I - I_2\|^2}{\frac{1}{k} + \frac{1}{n_2}} - \frac{\|I - I_1\|^2}{\frac{1}{k} - \frac{1}{n_1}}, \quad (3)$$

where the symbol $\|\ \|$ denotes an Euclidean distance.

$\Delta E_{merge}$ in (2) describes the increase of the total squared error caused by cluster merging that is treated as a reclassification of all data points from the cluster 1 into the cluster 2. As a result, the number of clusters is reduced by one. $\Delta E_{correct}$ in (3) describes the increment in the total squared error caused by partial reclassification of data points from the cluster 1 into the cluster 2, which doesn't influence on a number of clusters.

It is easy to verify that the difference between $\Delta E_{merge}$ and $\Delta E_{correct}$ is a perfect square:

$$\Delta E_{merge} - \Delta E_{correct} = \frac{\left\|\alpha \cdot (I - I_2) - \frac{I - I_1}{\alpha}\right\|^2}{\frac{1}{n_1} + \frac{1}{n_2}}. \quad (4)$$

Obviously the reclassification of data points provides a decrease of the total squared error in the case of the negative value of $\Delta E_{correct}$:

$$\Delta E_{correct} < 0. \quad (5)$$

Taking into account only the sign of the increment in the total squared error, we obtain from (3) and (5) the necessary condition for correction of the cluster 1:

$$\|I - I_1\| > \alpha \cdot \|I - I_2\|, \quad (6)$$

where the positive $\alpha$ is defined as:

$$\alpha \equiv \sqrt{\frac{n_2(n_1 - k)}{n_1(n_2 + k)}} < 1. \quad (7)$$

As the coefficient $\alpha$ decreases with increasing of $k$, then, all other things being equal, the reclassification of several data points is more efficient than the reclassification of individual data points. So, if the reclassification of one data point from the cluster 1 into the cluster 2 lowers the total squared error $E$, then the reclassification of remaining data points of the same real numbers from cluster 1 to cluster 2, reduces the error even further.

## 5. IDEA OF THE METHOD

The discussed formulas provide a clear alternative to the *K*–means method, referred to as *Kh*–method.

In our method, the reclassification of data point sets is performed without intermediaries like cluster centers. It is assumed that each cluster is divided into certain subsets, such as individual data points, or subsets of identical data points. In the simplest case, we consider the pairs of clusters, for example, all pairs, or some their subset, say, the pairs of adjacent segments in the context of image processing.

For optimal clustering the reclassification of sets of data points from one cluster into another according to the conditions of maximal minimization of the total squared error $\Delta E$ is produced. Among the ambiguous options to reduce the total squared error the option providing the greatest reduction is selected (Table).

Table.

**Comparison *K*–means with *Kh*–method**

|  | *K*–means | *Kh*–method |
|---|---|---|
| **Condition** | $\|I - I_1\| > \|I - I_2\|$ | $\|I - I_1\| > \alpha \cdot \|I - I_2\|$ |
| **Selecting of destination** | $\|I - I_2\| = \min$ | $\dfrac{\|I - I_2\|^2}{\frac{1}{k} + \frac{1}{n_2}} - \dfrac{\|I - I_1\|^2}{\frac{1}{k} - \frac{1}{n_1}} = \min$ |

The table explains our method in comparison with *K*–means [2–6]. The first line of the table describes the condition of starting or continuing of reclassification process, which is performed if the required cluster 2 exists. The second line describes the selection of the cluster 2 of several possible.

Similarly, as in the conventional *K–means* method, reduction of total squared error $E$ in *Kh–*method is performed iteratively. But our method results in stable clustering, which is defined formally and can't be improved by the total squared error $E$ using *K–means* method.

## 6. STABILITY CONDITION

Applying the logical negation to the expressions (5), (6), we obtain the stability condition:

$$\Delta E_{correct} \geq 0 \quad \Leftrightarrow \quad \|I - I_1\| \leq \alpha \cdot \|I - I_2\|. \quad (8)$$

**Definition.** A partition of data points into clusters is called stable iff reclassification of specified subsets of data points from one cluster into another does not reduce the total squared error $E$ for all considered pairs of clusters.

For example, in image processing [1], segmentation of an image is named stable with respect to reclassification of certain subsets of identical pixels from one segment to adjacent segment without damaging their integrity, iff any permissible reclassification causes an increasing of the total squared error $E$ or at least, leaves it unchanged.

Obviously, the optimal clustering is stable, but the stable clustering is not necessarily optimal. It should be noted that the partitions of the data points into clusters according to *K–means* method, in general case, is not stable, let alone optimal.

In *Kh*–method the split–and–merge techniques are used to generate the sequence of partitions of data points into different numbers of clusters, as in [6, 9–12]. At that in our method the stability of resulting partition along with reduction in the total squared error due to reclassification of subsets of data points from one cluster to another is provided by means of iterative correction of clusters. In contrast to the correction that is used to form a stable partition, the operation of cluster merging and also inverse operation of cluster splitting generally violate the stability of partitions. Therefore in *Kh*–method each violation of the stability, caused by cluster merging, is immediately compensated by means of iterative correction.

## 7. ADVANCED *Kh*– METHOD

For a fixed union of some considered clusters, an increment $\Delta E$ in the total squared error is an additive relative to composition of cluster modifications. So, in order to formalize exchanges between multiple clusters involving simultaneous reclassification of several sets of data points, the increase in the total squared error $\Delta E_{merge}$ caused by merging of considered clusters, is conveniently represented as:

$$\Delta E_{merge} = \Delta E + \Delta E'_{merge}, \quad (9)$$

where $\Delta E'_{merge}$ is an increase in the total squared error, caused by the merging of converted clusters originated of the same subset of data points.

In particular, for bijective converting of $l$ clusters into $l$ primed ones: $1 \rightarrow 1', 2 \rightarrow 2', \ldots l \rightarrow l'$, $\Delta E$ in (9) describes the increment in the total squared error $E$, caused by correcting of partitions that originated by specified subsets of data points: $1 \cup 2 \cup \ldots \cup l = 1' \cup 2' \cup \ldots \cup l'$. So in this particular case of interest, formula becomes:

$$\Delta E_{correct} = \Delta E_{merge} - \Delta E'_{merge}, \quad (10)$$

where a generalization of formula (2) for the case of merging of $l$ clusters is expressed as:

$$\Delta E_{merge} = \frac{\sum_{1 \leq i < j \leq l} n_i n_j \|I_i - I_j\|^2}{\sum_{1 \leq i \leq l} n_i}. \quad (11)$$

**Remark.** Obviously in (11) the summation $\sum_{1 \leq i < j \leq l}$ can be replaced by more familiar $\frac{1}{2}\sum_{i=1}^{l}\sum_{j=1}^{l}$, and $\sum_{1 \leq i \leq l}$ can be redesigned to $\sum_{i=1}^{l}$.

For illustrative example, one can check that in the case of $l = 2$, from (2) and (10) the expression (4) for $\Delta E_{merge} - \Delta E_{correct}$ is easily derived.

For the interpretation of (10), (11) it is important that the problem of minimizing of the total squared error $E$ is divided into the problems of maximizing of the functional $\Delta E_{merge}$, defined for certain overlapping tuples of clusters: pairs, triplets, quartets, etc.

So in our method an iterative correction starts with the modifying of pairs of clusters, which take away the data point subsets from each other to maximally decrease $E$ for the whole partition and concomitant increase of $\Delta E_{merge}$ for each pair of clusters under consideration. It is appropriate to once again repeat that the pairs of clusters, in general case, overlap each other.

After obtaining of the partition that is stable with respect to simple reclassification of any specified subset of data points from one cluster to another, the correction is performed for the next tuples of clusters. In this case, cluster triplets, quartets etc. are optimized owing to reclassifications of data points similarly to the previous tuples, so that stability strengthening is provided. As it follows from the construction, to reduce the total squared error $E$, the modification of each cluster at each step is necessary. Therefore in these stages of processing it is useful to treat synchronous cluster modifications by means of reclassification of several subsets of the data points, as in Otsu method [7, 8]. In the case of proper choice of the sets of data points that are to be reclassified, the total squared error $E$ effectively decreases with increasing of $l$, and then is stabilized closely to the required minimum for optimal partition, so that at a certain $l$ the partition of data points remains unchanged or varies within negligible limits.

It should be noted that the increase of tuples in the number of items often dramatically increases the processing time, which limits the total consideration of the tuples consisting of a large number of items. Therefore, it is not necessary to dismiss the task of constructing of the tuples with a variable number of items to optimize the conversion of each cluster.

In our method, the current stable partition is used as initial to generate the partition with the next number of clusters by split–and–merge techniques, starting from the trivial optimal partitions. As mentioned above, in the first of these partitions all points belong to a single cluster, and in the second partition each subset of identical data points, or each point in the case of conditional optimization, presents its own cluster. Thus the approximations of remaining optimal partitions can be produced one by one, either by splitting of some cluster from the current stable partition, either by merging of its clusters, or by utilizing both techniques.

Cluster merging or splitting is performed to provide a minimum of total squared error $E$ for the target partition, consisting of the number of clusters increased or decreased by one. To do this, the resultant total squared error $E$ is calculated for all admissible cluster conversions and the best merging or splitting, which corresponds to the minimal resultant error $E$, is carried out. To improve a minimizing of the total squared error $E$ its value is calculated taking into account the subsequent correction. So the merging criterion is expressed as:

$$\Delta E'_{merge} = \min, \qquad (12)$$

where $\Delta E'_{merge}$ is an increase in the total squared error, caused by the merging along with subsequent correction.

Apart from the constructing of tuples, the creative component of the method that affects the efficiency of the decision consists in a way to determine the subsets of data points, which are intended for reclassification from one cluster to another, which in the case of conditional optimization provides the formation of the clusters possessing the certain properties. The various implementation options of the method are to be developed and tested to choose the best for particular application conditions. As for the software, it should be noted that for proper implementation of the discussed type of optimization algorithms, an accurate calculation and online modification of current attributes of clusters is needed. In turn, online modification of cluster attributes entails the dependence of computational process on the order of treating of the data points. And just optimality guarantees invariant clustering results regardless of the ordering of the data points during scanning. Therefore a reliability of the calculated optimal or nearly optimal partitions should be the subject of close attention in addition to the usual substantive interpretation of the clustering.

## 8. EXPERIMENTAL RESEARCH

*Kh*–method has been worked out in the field of image processing where the task of creating of so–called "stable" algorithms providing invariant clustering (segmenting) of images regard-less of their variability and initial representation, such as the initial cluster centers in *K*–means method, is often treated [13]. In our method, the invariance is treated as a property of optimal partitions of an image, and the stability is defined as an attribute of the optimality. Thus, the problem reduces to finding of one or another algorithm for efficient computation of the sequence of optimal partitions[3].

*Kh*–method is the generalization of Otsu method and Mumford–Shah model.

In comparison with known solutions [7, 8], our software implementation of Otsu method is characterized in that the minimization of the total squared error $E$ is performed for overlapping parts of the histogram, and the increase of the total square error due to cluster merging $\Delta E_{merge}$ adjusted for subsequent correction is precomputed. The cluster tuples are defined by merging of consecutive ranges of intensity histogram, and subsets of pixels to be reclassified, are generated by means of exhaustive search of intensity thresholds.

Our software implementation of Mumford–Shah model differs from the known solutions, mainly in that, in addition to the traditional merging operation, it supports the correction of connected segments or clusters of non–connected pixels. The minimization of the total squared error $E$ in Mumford–Shah model is performed for overlapping parts of the image. In the discussed version of the software implementation, the correction is done inside the image parts covered by the tuples that are formed by each segment (cluster) together with the surrounding neighbors. In this case, the central donor cluster gives away the subsets of pixels into neighboring acceptor clusters from the condition of maximum decrease of the total squared error, which can be conveniently estimated by (3) or (10).

In our experience, modern computers are capable to provide a solution to the optimization problem and, at least the optimal or nearly optimal image approximations turns out to be available in image processing domain (Fig. 1).

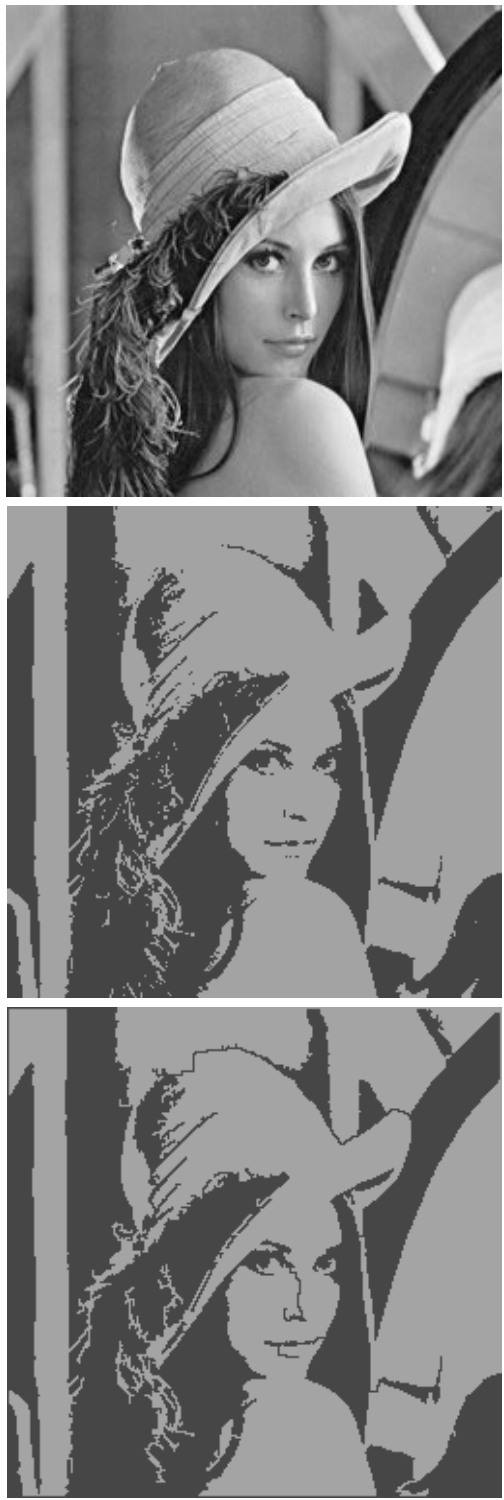

**Fig. 1:** The appearance of optimal partitions of the standard image (top) into two connected segments (bottom) and into two clusters of disconnected pixels (central).

Fig. 1 shows so–called piecewise constant approximations of the image obtained by substituting the source pixel intensities for their values averaged within the clusters and referred to in the same way as optimal or nearly optimal approximations. Piecewise constant approxima-tion in the middle of Fig. 1 is the well–known optimal binary representation of the standard image consisting of 351 connected segments. Just this approximation is obtained in the usual Otsu method [7] and can easily be found in electronic publications. The bottom approximation in Fig.1 consisting of two connected segments is so–called nearly optimal approximation that obtained in an automated interactive mode in the course of developing of *Kh*–method [1, 14]. The nearly optimal approximation has a clear structure and consists of areas connected by coupling elements one pixel wide. The coupling elements either naturally, either artificially fit into the image, such as the contour of the nose or fragments of the frame along the edge of the bottom image in Fig. 1. The discussed optimal and nearly optimal approximations look visually similar, and, apparently, the easiest way to get the second consists in converting the first.

---
[3] In this section one–dimensional clustering is under consideration.

A remarkable property of optimal approximations consists in detailed markings of visually perceived objects.

Numerically, the difference of an approximation from the image is described by the standard deviation that is estimated by the formula:

$$\sigma = \sqrt{\frac{E}{N}}, \quad (13)$$

where $E$ is the total squared error and $N$ is the number of pixels in the image. So, the above optimal and nearly optimal approximations are characterized by the values $\sigma = 30{,}64564$ and $\sigma = 31{,}60341$, respectively.

The overall results of segmentation according to *Kh*–method are presented graphically in Fig. 2, which demonstrates the dependencies of the standard deviation $\sigma$ on the cluster number in the range from 1 to 1000.

In Fig. 2 the central dashed curve and the curves over it describe the sequence of partitions into connected segments and demonstrate the dependence of the standard deviation of the number of segments. Two bottom curves that are shown as a function of the cluster number, describe the sequence of data partitions into the clusters of disconnected pixels. As well as the central dashed curve, the bottom two curves correspond to the sequence of overlapping partitions, while the other curves describe the hierarchies of partitions.

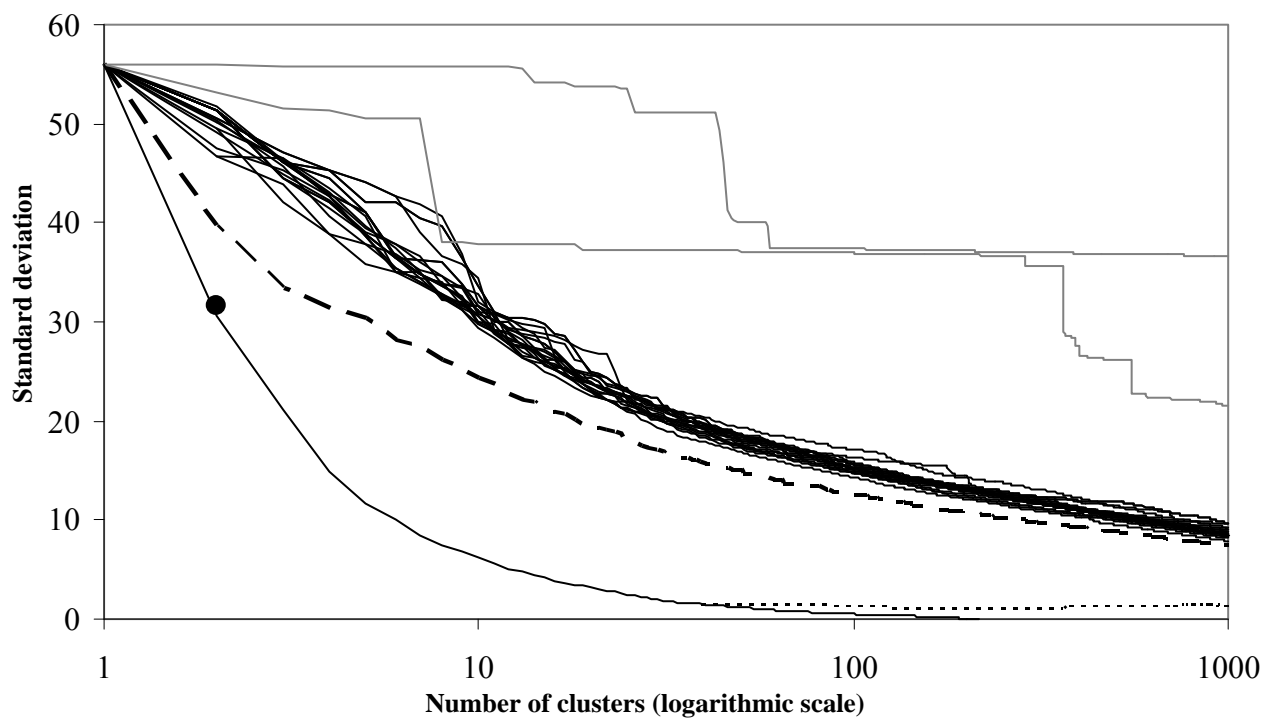

**Fig. 2:** The dependencies of standard deviation $\sigma$ on cluster number.

The values of the standard deviation $\sigma$ for the optimal piecewise constant approximations of the image with a different number of averaged intensities were obtained in our generalized version of multi–threshold Otsu method. These values are described by drop–down left curve, which doesn't depend on any rearrangement of pixels within the image and bounds the values of $\sigma$, attainable for the corresponding number of clusters. The bold black dot, located just above this curve, describes the nearly optimal approximation of an image consisting of two connected segments (bottom in Fig. 1).

The upper two gray curves were obtained for the heuristic merging of connected segments with a minimum difference of averaged intensities. They describe the old–fashioned segmentation that destroys the image especially in the case of small number of segments. The next twenty interwoven black curves below were obtained for the variants of the criterion (2) of segment merging derived in Mumford–Shah model and its versions [9-12, 14]. For a small number of segments these criteria provide a detection of some large objects. However, the number of objects is missing, and, moreover, the variability of an image affects the detection of specific objects.

The central dashed curve describes the improved approximations, which are not hierarchical. This curve is obtained according to the version of *Kh*–method, based on Mumford–Shah model, and target approximations are produced by means of correcting (3) of connected segments performed in turn with their merging. In this case, the conditional optimization is curried out, so that the reclassification of pixels, violating the connectivity of segments is locked.

As seen in Fig. 2, the correction of connected segments provides a significant improvement of the standard deviation, which drops down for all cluster numbers, since the correction compensates the violation of the stability and accumulation of systematic errors caused by segment merging.

This increases the number of detected objects and reduces the number of missing ones. However, concerning the optimality note that in the case of perfect approximations the central dashed curve in the Fig. 2 should pass through the bold dot. But it lies above. Then, though a noticeable improvement of image approximations with connected segments is achieved, its capabilities have not yet been exhausted.

For further improvement of the approximations of the image, we have developed a version of their computing without support of connectivity of pixels within the treated sets. In this case, the connectivity of pixels within the cluster is not taken into account, and the merging of adjacent clusters with neighboring pixels from both clusters is only treated. It turned out that this provides a close reproduction of the optimal partitions, and if the number of clusters is less than 30–40 the corresponding curve (bottom, dashed line in Fig. 2) merges into the curve for optimal partitions. This close coincidence of the curves indicates the expressed minimums of standard deviation $\sigma$ as well as total squared error $E$. It also enables to suppose that the condition of connectivity of pixels within the segments is the only obstacle to really get the optimal partitions. Likely, for the traditional object detection in terms of connected segments the intermediate solution based on weakening of connectivity of pixels may prove to be effective, if the weakened connectivity is determined by the enlarged (non–minimal) threshold of geometric distance.

For motivation of the computing of optimal approximations it is important that the sequence of optimal partitions of the image for the first tens of cluster numbers are insignificantly altered by reducing the size of the image. In this case the dependence of the standard deviation on the number of clusters cozies from above with the curve for optimal partitions and provides a numerical estimation for the deviations from the invariant segmentation. In a more general sense it is quite possible that the optimal approximations are the simplest means to ensure an invariant segmentation for variable images. If so, then the invariant segmentation is provided, not due to dodgy algorithm, but owing to effective solution of formal optimization problem Just the requirement of invariance seems sufficient for solving the so–called "segmentation problem" in image processing domain. A disclosure of invariant segmentation in terms of visually perceived "objects" seems the secondary one, which is essential for simulation of only human visual perception.

The development of a multidimensional version of our method is not labor–consuming and seems attractive for the segmentation of color and multispectral images. Slightly more complicated is the study of features of pixels that are characterized by the consecutive differences of their intensities in the image and in each optimal approximation. As for the hardware resources, conventional computer memory is more than enough to compute the optimal image approximations. But currently, software implementation of *Kh*–method faces the computational speed problem. At the same time, to avoid excessive efforts for standard optimization of computations, it makes sense to parallelize the algorithms on a suitable multiprocessor system.

## 9. CONCLUSION

Thus, in the paper, we have proposed the *Kh*–method for clustering of multidimensional data points, and illustrated one–dimensional version of this method by the example of digital image segmentation.

We suggest that the target optimal partitions have the required property of invariance, and fluctuate within acceptable limits depending on the variability of the input data. Stability is regarded as a feature of optimality, so that we define the stability of clustering by analogy with the equilibrium stability of the mechanical system. If the stability of the simplest mechanical system is described by the minimum of potential energy, then for clustering it is associated with the minimum of the total squared error.

Our method is focused on obtaining of complete sequence of the optimal partitions for all numbers of clusters. Starting from the trivial optimal partitions, we obtain the remaining by means of known split–and–merge techniques. At the same time, cluster merging or splitting in our method is alternated with correction, which converts the partitions into stable ones.

In contrast to heuristic *K*–means method, the correction in our method is analytically justified, so that *Kh*–method provides to reduce the resultant value of total squared error, because overcomes the premature interruption of minimization process, utilizes more accurate selection of reclassification option of several possible, and envisages the reclassification of certain sets of data points. Cluster partitioning into subsets of data points to be reclassified contributes to the reduction of the total squared error. In turn, the clusters themselves are combined into overlapping tuples, which are iteratively processed by means of portionwise optimizing of the partitions of data point set.

It should be noted that for uncomplicated computing in available amount of RAM, it is advisable to use the special data structure. In our experience, the easiest way is to develop of appropriate software in terms of Sleator–Tarjan dynamic trees, which in modern notation are referred to as "Disjoint sets", "Persistent data structures" or "Splay trees" [15-17]. To avoid a tedious reasoning, in this paper we have omitted details of our method software implementation. The minimum number of algorithms required for confident use of Sleator–Tarjan dynamic trees, which has been worked out in image processing domain, will be presented in the next paper.